\documentclass[11pt]{article}
\usepackage[margin=2cm]{geometry}
\usepackage[T1]{fontenc}
\usepackage[utf8]{inputenc}
\usepackage{lmodern}
\usepackage{graphicx}
\usepackage[hidelinks]{hyperref}
\usepackage{microtype}
\usepackage{setspace}
\let\SkeletonOriginalSection\section
\RenewDocumentCommand{\section}{s o m}{%
  \IfBooleanTF{#1}{\SkeletonOriginalSection*{#3}}{\IfNoValueTF{#2}{\SkeletonOriginalSection{#3}}{\SkeletonOriginalSection[#2]{#3}}}%
  \mbox{}\par
}
\let\SkeletonOriginalSubsection\subsection
\RenewDocumentCommand{\subsection}{s o m}{%
  \IfBooleanTF{#1}{\SkeletonOriginalSubsection*{#3}}{\IfNoValueTF{#2}{\SkeletonOriginalSubsection{#3}}{\SkeletonOriginalSubsection[#2]{#3}}}%
  \mbox{}\par
}

\title{TEMPLAR Wales: A georeferenced environmental and toponymic dataset of Welsh settlements}
\author{Oktay Karakuş\textsuperscript{1} \and Can Eyüpoğlu\textsuperscript{1,2}}
\date{\textsuperscript{1}Department of Computer Science and Informatics, Cardiff University, Cardiff CF24 4AG, UK\\
\textsuperscript{2}Department of Computer Engineering, Turkish Air Force Academy, National Defence University, Istanbul 34149, Türkiye\\[0.5em]
Corresponding author: Oktay Karakuş (karakuso@cardiff.ac.uk)}

\begin{document}
\raggedbottom
\maketitle

\begin{abstract}

Place names provide persistent records of how landscapes have been described
and organised, but their quantitative reuse requires explicit separation
between mapped places, lexical annotations and environmental measurements.
TEMPLAR Wales is a georeferenced environmental-toponymy dataset comprising
3,757 settlement records across Wales. The resource links a reproducible
settlement frame to deterministic lexical screening and settlement-level
environmental attributes through stable identifiers. It contains 1,350 lexical
detections across 1,294 settlements, generated from a frozen registry of 24
Welsh place-name elements, while retaining exact- and prefix-token matches and
their provenance separately. Environmental attributes describe river and
coastal proximity, elevation and local terrain context at multiple spatial
scales, land cover and neighbourhood woody cover, with parallel terrain
measurements derived from independent elevation products. The dataset is
distributed as four relational tables accompanied by a field-level data
dictionary, source-provenance register and licensing metadata. Technical
validation confirms relational integrity, deterministic lexical reconstruction,
documented environmental coverage, strong agreement between independent terrain
sources and reproducible reconstruction of the frozen release. TEMPLAR Wales
provides a reusable foundation for research in toponymy, linguistic geography,
historical and environmental landscape studies, GIS and spatial data analysis
without treating computational lexical detections as verified etymologies or
contemporary environmental measurements as historical landscape
reconstructions.

\end{abstract}

\section{Background \& Summary}

Place names constitute a persistent form of cultural and geographic
information. They encode ways in which landscapes have been described,
distinguished and organised, while remaining visible in maps, gazetteers,
administrative systems and everyday geographic reference long after the
circumstances in which individual names originated may have changed
\cite{RiescoChueca2010,Reszegi2020,Williamson2023}. Toponymic information has
consequently been used across historical geography, landscape research and
cultural-environmental studies to investigate relationships between naming,
environmental characteristics and landscape change
\cite{Conedera2007,CalvoIglesias2012,FagundezIzco2016,Atik2022,Hearn2024}.
At the same time, place names are culturally and linguistically mediated
records rather than direct environmental observations. Their interpretation
can depend on historical forms, language, morphology, local usage and
documentary evidence, making transparent representation of both lexical
evidence and its limitations essential for quantitative reuse
\cite{Reszegi2020,Williamson2023}.

Wales provides a particularly informative setting for constructing such a
resource. Welsh settlement names occur within a geographically varied
landscape, while the naming system itself reflects layered linguistic and
historical processes. Welsh place-name scholarship emphasises the importance
of historical forms, linguistic components and contextual interpretation,
rather than assuming that contemporary written forms can be interpreted
unambiguously from their surface appearance \cite{Owen2015,Parry2023}. These
characteristics make Wales valuable for environmental-toponymy research, but
they also expose a broader data problem: reproducible quantitative analysis
requires a clear distinction between the mapped settlement record, the name
selected for analysis, a computational lexical detection, the documented
interpretation of a registered lexical element and the environmental
measurements associated with the settlement location.

Existing studies demonstrate the potential of toponymic information for
environmental and landscape research. Place names have been used to reconstruct
past land use and disappearing landscape features
\cite{Conedera2007,CalvoIglesias2012}, to investigate associations between
plant-related names and environmental and social conditions
\cite{FagundezIzco2016}, to characterise rural and agrosilvopastoral landscapes
\cite{Atik2022,Hearn2024}, and to examine relationships between toponymic
diversity, vegetation and landscape characteristics
\cite{Valko2023,Zhou2025}. Recent work has further demonstrated the increasing
role of GIS and geostatistical approaches in connecting toponymic information
with spatially explicit landscape evidence
\cite{Fuchs2015,Guo2025,MitxelenaHoyos2026}. Such studies establish the value
of toponymic information as a research source, but the datasets underpinning
individual analyses are commonly assembled around particular questions.
Reusable resources that explicitly separate settlement observations, lexical
annotations, environmental measurements and their respective provenance can
therefore support broader comparative and methodological applications.

TEMPLAR Wales addresses this need by providing a versioned, georeferenced
environmental-toponymy dataset for 3,757 settlement records across Wales. The
resource integrates three conceptually distinct information layers. First, a
settlement layer preserves mapped settlement identity, geographic location,
source naming fields and the provenance of a deterministically selected
analytical name, derived from OS Open Names \cite{OSOpenNames2026}. Second, a
lexical layer applies a frozen deterministic detector to those analytical names
using a source-audited registry of 24 Welsh place-name elements. Third, an
environmental layer describes hydrological, coastal, terrain and land-cover
characteristics at the settlement locations using multiple geospatial products
\cite{OSOpenRivers2025,OSTerrain502025,OSBoundaryLine2026,
CopernicusCLMS2018,CopernicusGLO90}. These layers are linked through stable
project identifiers while remaining available as separate machine-readable
records.

A central design principle of the resource is separation between observation,
annotation and interpretation. The settlement frame represents mapped source
records rather than unique written names. The analytical-name field records the
single source name selected under a frozen processing rule rather than asserting
a definitive linguistic classification. Lexical detections record reproducible
string-level matches rather than validated individual-name etymologies, an
important distinction given the historical and linguistic complexity of Welsh
toponymy \cite{Owen2015,Parry2023}. Environmental variables describe
contemporary or product-specific geographic conditions rather than reconstructed
historical environments. These distinctions are represented explicitly through
provenance, language-status, match-type, quality, coverage and eligibility
fields rather than being left as implicit assumptions of a downstream
analysis.

The released lexical component contains 1,350 detections across 1,294
settlements and retains exact- and prefix-token matches separately. Settlements
without a registered detection remain part of the complete settlement frame,
allowing users to define lexical contrasts appropriate to their own research
questions rather than receiving a prefiltered analytical sample. The associated
24-element registry records canonical forms, concise project-authored glosses,
semantic classifications, source identifiers and locators, and explicit
limitations on interpretation. The registry is therefore both a computational
input and a provenance record for the lexical annotations.

The environmental component provides settlement-level measurements derived
from OS Open Rivers \cite{OSOpenRivers2025}, OS Terrain 50
\cite{OSTerrain502025}, OS Boundary-Line \cite{OSBoundaryLine2026}, CORINE
Land Cover 2018 \cite{CopernicusCLMS2018} and Copernicus DEM GLO-90
\cite{CopernicusGLO90}. Core river, coastal, terrain and point-level land-cover
attributes are represented across the complete settlement frame, while
neighbourhood woody-cover measurements retain explicit coverage and eligibility
information where measurements are unavailable. Terrain is represented at
multiple neighbourhood scales, and independently derived OS Terrain 50 and
Copernicus GLO-90 measurements are included to support source-sensitivity
assessment. Raw provider geometries and raster products are not redistributed;
the resource instead provides documented settlement-level observations and
derived attributes together with field-level provenance.

The public release is organised around four principal records:
\texttt{settlements.csv}, \texttt{lexical\_detections.csv},
\texttt{environmental\_attributes.csv} and
\texttt{lexical\_registry.csv}. A field-level data dictionary, source
provenance register, licensing documentation, machine-readable manifest and
cryptographic checksum inventory accompany these tables. The dataset is
generated through a deterministic export procedure with automated checks for
identifier integrity, table relationships, expected schema, lexical
reconstruction, environmental coverage and exclusion of non-release material.
This structure is intended to make the resource inspectable and reusable
without requiring access to the research workflow from which it originated.

TEMPLAR Wales can support several classes of reuse. Researchers can examine the
spatial distribution of registered place-name elements, develop or benchmark
alternative lexical-detection approaches, link settlement naming to additional
environmental or historical information, investigate geographic variation in
toponymic composition, or use the environmental measurements independently of
the lexical annotations. Such applications extend established uses of
toponymic evidence in historical-landscape reconstruction, environmental
interpretation and spatial landscape analysis
\cite{CalvoIglesias2012,FagundezIzco2016,Atik2022,Hearn2024,
Valko2023,Zhou2025}. The stable settlement frame also provides a basis for
enrichment with historical name forms, archival maps, geology, soils,
hydrological or climatic information, historical land cover and independently
curated linguistic annotations. Because the lexical detections and
environmental attributes remain separate relational layers, extensions need not
adopt the analytical categories or statistical design of the study that
motivated the original dataset.

The resource was initially developed in connection with a preregistered study
of environmental information in Welsh settlement names
\cite{Karakus2026Preregistration}. The dataset released here is deliberately
broader than the inferential outputs of that application. It excludes
coefficients, hypothesis-test results, model predictions, residuals and
cross-validation outputs, retaining instead the reusable observations,
annotations, environmental attributes and provenance from which alternative
analyses can be constructed. The associated analytical study therefore
represents one application of the resource rather than defining the limits of
its use.

By releasing the settlement, lexical and environmental layers together with
their construction rules and interpretative boundaries, TEMPLAR Wales provides
a reusable bridge between traditional place-name evidence and computational
spatial analysis. Its purpose is not to reduce Welsh toponymy to automated
string matching or to treat contemporary environmental data as a direct record
of naming history. Rather, it provides a transparent data structure in which
those different forms of evidence can be linked, inspected and extended while
their provenance and limitations remain explicit.

\section{Methods}

The TEMPLAR Wales dataset was constructed through a staged workflow that
integrates settlement gazetteer information, reproducible lexical annotation
and multiple contemporary environmental data sources. The workflow begins by
defining a Wales-wide settlement frame and selecting a single analytical name
for each retained record. A frozen, source-audited lexical registry is then
applied to these names using a deterministic detection procedure. In parallel,
settlement-level hydrological, coastal, terrain and land-cover attributes are
derived from independent geospatial source products. These components are
finally integrated through stable project identifiers into the four relational
data records described in this Data Descriptor.

The following subsections document this construction from source data to public
release. We first define the scope and relational design of the dataset and
describe the upstream geographic and lexical sources. We then detail
construction of the settlement frame and analytical names, followed by the
lexical registry and ETDE detection procedure. The environmental-data workflow
is described separately for river and coastal proximity, OS Terrain 50-derived
terrain attributes, independent Copernicus DEM GLO-90 terrain measurements,
and CORINE-derived land-cover and woody-cover attributes. The final subsection
describes integration of these components, identifier relationships, deterministic
export and dataset versioning. Together, these stages define the complete
provenance chain from upstream records and source products to the released
settlement-level resource.

\subsection{Dataset design and scope}

The dataset was designed as a settlement-level resource linking contemporary
geographic and environmental attributes to reproducible lexical annotations of
Welsh place names. The observational frame comprises 3,757 mapped settlement
records in Wales. Each retained settlement is represented by a stable
project-level identifier and can be linked to its naming metadata, deterministic
lexical detections and environmental attributes.

The resource was constructed as four related data records rather than as a
single analysis table. \texttt{settlements.csv} defines the settlement frame
and analytical names; \texttt{lexical\_detections.csv} stores long-format
lexical detections; \texttt{environmental\_attributes.csv} contains
settlement-level environmental measurements and quality fields; and
\texttt{lexical\_registry.csv} documents the frozen lexical elements used by
the detection procedure. This separation preserves the distinction between
source identity, derived lexical annotation and environmental measurement, and
allows each component to be reused independently.

The released dataset contains observational and deterministically derived
attributes only. Inferential statistics, fitted model quantities, residuals,
permutation results, geographically held-out predictions and other outputs
generated in subsequent analyses are not included.

\subsection{Source datasets}

Settlement locations and source naming fields were obtained from Ordnance Survey
(OS) Open Names \cite{OSOpenNames2026}. Environmental attributes were derived from OS Open Rivers
\cite{OSOpenRivers2025}, OS Terrain 50 \cite{OSTerrain502025}, OS Boundary-Line High Water Mark
\cite{OSBoundaryLine2026}, CORINE Land Cover 2018 \cite{CopernicusCLMS2018} and
Copernicus DEM GLO-90 \cite{CopernicusGLO90}. The released data contain settlement-level source fields
or derived measurements rather than redistributed copies of the underlying
provider geometries, vector networks or raster products.

OS Open Names provided the source settlement records, names, language metadata,
settlement classifications and mapped point locations. The retained local
snapshot records an embedded source timestamp of 29 January 2026. OS Open
Rivers supplied the mapped river network used to derive settlement-to-river
distance, with the retained local product carrying an October 2025 timestamp.
Elevation and neighbourhood terrain attributes were derived from the May 2025
OS Terrain 50 package. Coastal distance was derived from the High Water Mark
contained in the May 2026 OS Boundary-Line product.

Land-cover information was obtained from the 2018 CORINE Land Cover product
(\texttt{U2018\_CLC2018\_V2020\_20u1}), and an independent terrain measurement
layer was derived from Copernicus DEM GLO-90. Source-specific versions,
snapshots, provenance qualifications, transformations and licensing conditions
are recorded in \texttt{source\_provenance.csv} and
\texttt{LICENSES.md}.

The lexical resource was constructed from a source-audited set of Welsh
place-name elements. Lexical-source alignment used the Royal Commission on the
Ancient and Historical Monuments of Wales (RCAHMW) Historic Place Names of
Wales resources and Geiriadur Prifysgol Cymru Online (GPC)
\cite{RCAHMWPlaceNames,GPCOnline}. The released
registry contains concise project-authored summaries and source locators rather
than reproduced dictionary definitions.

\subsection{Settlement-frame construction}

The settlement frame was constructed from OS Open Names point records located
in Wales. Records were retained when their \texttt{local\_type} belonged to
one of five settlement categories: \texttt{Village}, \texttt{Hamlet},
\texttt{Suburban Area}, \texttt{Other Settlement} or \texttt{Town}. The
resulting frame contains 3,757 source records: 1,361 villages, 1,198 hamlets,
866 suburban areas, 190 other settlements and 142 towns.

The unit of observation is an upstream settlement record at a mapped point
location, not a unique written place name. Settlements sharing the same
normalised name were therefore retained as separate observations when they
represented distinct source records. Similarly, coincident coordinates were
not automatically deduplicated. The final frame contains 3,328 normalised
analytical-name groups and 3,755 coordinate clusters. Four records occur as two
pairs of coincident point locations and remain in the resource as distinct
upstream records.

Each settlement was assigned a deterministic project identifier,
\texttt{templar\_id}. The upstream OS identifier is retained separately as
\texttt{os\_open\_names\_id} to support provenance and linkage to the source
product. The project identifier is used as the primary relational key across
the released settlement, lexical-detection and environmental tables.

Source coordinates were retained in the British National Grid coordinate
reference system (EPSG:27700). Geographic latitude and longitude were derived
through the documented coordinate transformation and are supplied as additional
reuse-oriented fields rather than as replacements for the source projected
coordinates.

\subsection{Analytical-name construction}

A single analytical name was selected deterministically for each settlement
before lexical detection. The purpose of this field is to provide a stable,
reproducible string against which the lexical detector can operate; it is not
intended to resolve the complete linguistic or historical identity of an
individual place name \cite{Owen2015,Parry2023}.

Where the available Open Names language metadata explicitly identified an
appropriate Welsh-labelled source name, that name was selected according to the
frozen name-selection rule. Otherwise, the primary Open Names name field was
retained and its analytical language status was recorded as unresolved. This
procedure selected the primary \texttt{name1} field for 3,618 records, of
which 203 carried an explicit Welsh label and 3,415 had unresolved analytical
language status. For 139 records, a Welsh-labelled \texttt{name2} field was
selected instead.

Every settlement therefore has a non-missing analytical name and an explicit
record of the source field from which that name was selected. The released
metadata distinguish an unresolved language status from a positive
classification into another language.

For reproducible matching and grouping, analytical names were additionally
normalised using the frozen lexical-processing convention. Text was converted
to lowercase, Unicode accents and diacritics were normalised through NFKD
decomposition, hyphens and apostrophes were treated as token boundaries, and
characters outside \texttt{a--z} were removed from the matching representation.
The original selected analytical string remains available alongside its
normalised representation and associated grouping fields.

\subsection{Lexical resource}

The lexical component comprises a frozen registry of 24 Welsh place-name
elements selected and documented before construction of the released detection
table. Each registry entry has a canonical form, concise project-authored gloss,
semantic classification, source identifier and locator, and an explicit
statement of interpretative alignment or limitation.

Where an element had a predefined role in the associated environmental-toponymy
study, that role is retained as provenance metadata in the registry. This field
records the historical analytical design of the resource; it does not encode
the outcome of a statistical test and should not be interpreted as evidence
that the corresponding environmental relationship holds for an individual
settlement.

The registry was source-audited against RCAHMW and GPC resources
\cite{RCAHMWPlaceNames,GPCOnline}. The objective
was not to create a general Welsh etymological dictionary, but to define a
small, versioned lexical system that could be applied reproducibly to the
settlement frame. The released source fields therefore identify where the
element-level alignment can be checked while avoiding reproduction of
substantial source definitions.

Registry version information is retained in the released table so that future
extensions can add or revise lexical elements without silently changing the
meaning of detections produced by the frozen version.

\subsection{Lexical detection}

Lexical annotation was performed using the frozen ETDE v1 deterministic
detection procedure. The detector operates on the normalised analytical name
rather than on all available source-name variants, ensuring that each settlement
is processed using the same predeclared name-selection and normalisation
sequence.

Each analytical name was tokenised after normalisation. Registry elements were
then evaluated using the frozen exact-token and prefix-token matching rules.
Exact matches identify tokens equal to a registered element, whereas prefix
matches retain cases satisfying the corresponding ETDE prefix rule. Match type
is preserved in the released detection table rather than collapsing both forms
into a single binary lexical indicator.

The complete procedure was applied to all 3,757 analytical names. The resulting
long-format table contains one row per detected element--settlement relation and
retains the matched token, token position, match type, canonical registry
element, analytical-name source and relevant detector and registry versions.

A lexical detection establishes only that the frozen string-processing rule
identified a registered form in the analytical name. It does not establish the
historical etymology of the settlement, perform morphological parsing, determine
language identity, or demonstrate that the detected element historically
referred to the contemporary environmental attribute associated with its
semantic category. These distinctions are retained explicitly in the dataset
documentation to support appropriate downstream reuse.

\subsection{River and coastal attributes}

Hydrological context was represented by distance from each settlement point to
the nearest mapped feature in OS Open Rivers. Distance was calculated in the
projected British National Grid coordinate system so that the released
settlement-level value is expressed as a metric spatial measurement. The
underlying river network is not redistributed with the dataset.

The released river-distance field should be interpreted relative to the
representation and generalisation of the source network. It measures proximity
to the nearest feature represented in the retained Open Rivers product and is
not a measure of historical hydrology, channel permanence, discharge or
catchment membership.

Coastal context was represented by distance from the settlement point to the
High Water Mark contained in OS Boundary-Line. The calculation was likewise
performed in the projected coordinate system. Only the derived
settlement-level distance and relevant provenance or validity information are
released; the source Boundary-Line geometry is excluded from the dataset.

\subsection{Terrain attributes}

The principal terrain attributes were derived from OS Terrain 50. For each
settlement, the resource contains point elevation and neighbourhood terrain
summaries calculated at 1-km, 2-km and 5-km spatial scales.

For a settlement with elevation $z_i$, local terrain position at neighbourhood
scale $r$ was represented as the difference between the settlement elevation
and the mean valid terrain elevation within the corresponding neighbourhood:

\begin{equation}
T_{i,r} = z_i - \bar{z}_{i,r},
\end{equation}

where $\bar{z}_{i,r}$ is the mean terrain elevation of raster-cell centres
within a circular buffer of radius $r$ metres centred on the settlement point,
as defined by the frozen processing workflow. Positive values indicate that the
settlement point lies above its surrounding terrain reference and negative
values indicate that it lies below that reference. The dataset retains the
underlying point elevation, neighbourhood means, local terrain-position values
and associated validity fields rather than reducing the terrain information to a
categorical label.

The 1-km, 2-km and 5-km attributes were generated using the same frozen
construction procedure. These multiple scales are supplied to permit
scale-aware reuse and sensitivity analysis; none should be interpreted as a
universally preferred neighbourhood definition.

Raw Terrain 50 tiles and derived raster mosaics are not redistributed. The
released data contain only settlement-level derived measurements and provenance
information.

\subsection{Independent Copernicus terrain attributes}

A second terrain layer was generated from Copernicus DEM GLO-90 to provide an
independently sourced elevation-based measurement for reuse and technical
validation. OS Terrain 50 provides the principal terrain representation, whereas
Copernicus DEM GLO-90 is distributed as a digital surface model rather than a
bare-earth digital terrain model; it was used as an independent elevation surface
to assess the source sensitivity of the derived settlement-level terrain
attributes, rather than as an interchangeable product or ground-truth
representation. GLO-90-derived attributes were
calculated for all 3,757 settlements using a 2-km neighbourhood construction
corresponding to the principal settlement-level terrain representation.

The released fields include the Copernicus-derived settlement elevation,
neighbourhood terrain reference and local terrain-position measure, together
with validity information required to interpret the derived values. The
Copernicus fields are retained alongside, rather than substituted for, the OS
Terrain 50 attributes because the two products differ in source, spatial
resolution and production lineage.

Raw GLO-90 tiles and mosaics are not included in the release. Product-specific
provenance, acknowledgement and licensing information is provided in the
supporting metadata.

\subsection{Land-cover and woody-cover attributes}

Contemporary land-cover attributes were derived from CORINE Land Cover 2018.
Each settlement was associated with the corresponding point-level CORINE class
where the source coverage and assignment rules were satisfied. The dataset
retains the resulting class information together with quality or assignment
fields needed to distinguish substantive categories from processing or coverage
conditions.

Neighbourhood woody-cover attributes were additionally constructed at 500-m,
1-km and 2-km scales. The released variables describe the fractions of eligible
neighbourhood area assigned to the frozen forest, shrub or combined woody-cover
definitions, together with the corresponding coverage and eligibility
information.

Missing woody-cover values were retained where the frozen coverage criteria
were not satisfied. They were not replaced with zero, because absence of an
eligible neighbourhood measurement is distinct from an observed woody-cover
fraction of zero. Consequently, downstream users should evaluate the associated
coverage and eligibility fields when selecting records for analyses involving
these variables.

The source CORINE polygons are not redistributed. Only settlement-level class
assignments, derived neighbourhood fractions and associated quality fields are
included in the public resource.

\subsection{Dataset integration and identifiers}

The four scientific tables were assembled through explicit relational keys.
\texttt{templar\_id} uniquely identifies a settlement in
\texttt{settlements.csv} and occurs once for the corresponding record in
\texttt{environmental\_attributes.csv}. The same identifier provides the
one-to-many relationship to \texttt{lexical\_detections.csv}, because a
settlement can contain zero, one or more registered lexical detections.

Lexical detections are linked to \texttt{lexical\_registry.csv} through the
canonical element field. Every released detection is required to resolve to
exactly one entry in the frozen registry. Settlements without a registered
lexical detection remain present in the settlement and environmental tables and
have no corresponding row in the long-format detection table.

The public export was generated using a deterministic build procedure from the
frozen project data products. Export validation checks enforce identifier
uniqueness, relational integrity, expected table dimensions, categorical
domains, required-field completeness, field-specific missing-value semantics
and exclusion of non-release material. The released tables are accompanied by
a field-level data dictionary, source-provenance register, licensing
documentation, release manifest and checksum inventory.

Versioning is applied at the dataset level as well as to the lexical registry
and detector. The initial public resource is designated version 1.0.0. This
versioning structure is intended to allow future additions or corrections to be
distinguished from the frozen data records described here.

\section{Data Records}

The TEMPLAR Wales dataset is organised as a relational, machine-readable
resource centred on 3,757 settlement records in Wales. The published version
1.0.0 dataset is archived on Zenodo \cite{KarakusEyupoglu2026TEMPLAR}. The dataset separates
settlement identity and naming information, lexical detections, environmental
attributes and lexical-registry metadata into four primary tables linked by
stable project identifiers. Supporting files provide field-level definitions,
source provenance, licensing information, release metadata and integrity
checks.

The primary relational key is \texttt{templar\_id}, a deterministic
project-level identifier assigned to each settlement record. The upstream
Ordnance Survey Open Names identifier is retained as
\texttt{os\_open\_names\_id} for traceability, but is not used as the primary
identifier of the integrated resource. This design allows the dataset to retain
a stable internal relational structure while preserving linkage to the source
settlement record.

The dataset is distributed as four scientific tables:
\begin{itemize}
\item \texttt{settlements.csv}, 
\item \texttt{lexical\_detections.csv},
\item \texttt{environmental\_attributes.csv} and
\item \texttt{lexical\_registry.csv}. 
\end{itemize}

These are accompanied by
\begin{itemize}
\item \texttt{data\_dictionary.csv}, 
\item \texttt{source\_provenance.csv},
\item \texttt{README.md}, 
\item \texttt{LICENSES.md}, 
\item \texttt{manifest.json} and
\item \texttt{SHA256SUMS.txt}. 
\end{itemize}

\begin{figure}[htbp]
\centering
\includegraphics[width=\textwidth]{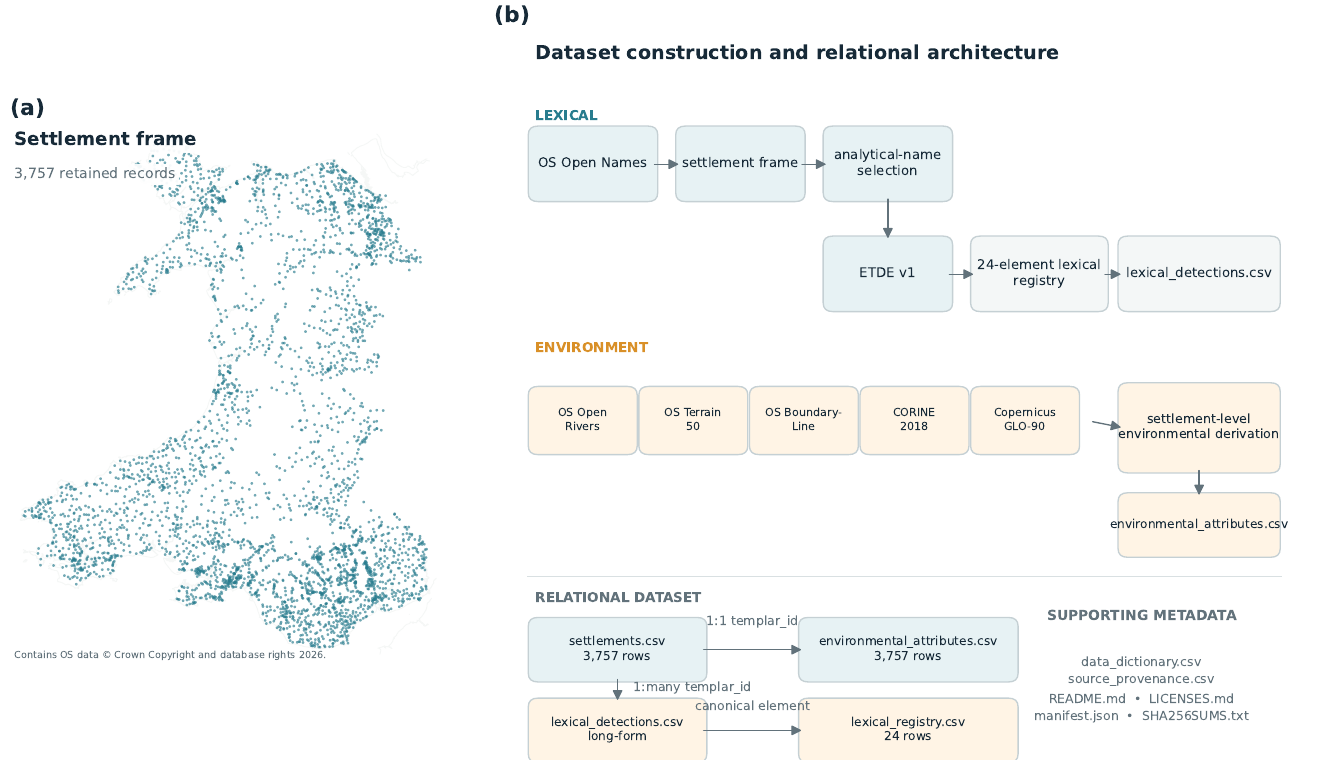}
\caption{\textbf{Dataset overview and construction architecture.}
\textbf{(a)} Geographic distribution of the 3,757 retained Welsh settlement
records. \textbf{(b)} The lexical stream applies analytical-name selection,
the 24-element lexical registry and ETDE v1 to generate
\texttt{lexical\_detections.csv}; the environmental stream derives
settlement-level attributes from OS Open Rivers, OS Terrain 50, OS
Boundary-Line, CORINE 2018 and Copernicus GLO-90. The lower portion shows the
four released scientific tables and their principal \texttt{templar\_id} and
canonical-element relationships. Supporting metadata provide field definitions,
provenance, licensing and release-integrity information. Contains OS data
\textcopyright\ Crown Copyright and database rights 2026.}
\label{fig:dataset_overview}
\end{figure}

Table~\ref{tab:data-records} summarises the principal records. An overview of
the settlement frame, construction streams and relational dataset architecture
is provided in Fig.~\ref{fig:dataset_overview}.

\begin{table}[htbp]
\centering
\caption{Primary data records in the TEMPLAR Wales dataset.}
\label{tab:data-records}
\begin{tabular}{p{0.36\textwidth} p{0.07\textwidth} p{0.07\textwidth} p{0.42\textwidth}}
\hline
\textbf{File} & \textbf{Rows} & \textbf{Columns} & \textbf{Content} \\
\hline
\texttt{settlements.csv}
& 3,757
& 23
& Settlement identity, source identifiers, original and analytical name fields,
language-status metadata, settlement type, British National Grid coordinates,
derived geographic coordinates and grouping fields for repeated names and
coincident locations. \\

\texttt{lexical\_detections.csv}
& 1,350
& 13
& Long-format deterministic ETDE v1 lexical detections linked to settlement
records, including canonical element, matched token, token position, match
type, semantic classification, preregistered role and analytical-name source. \\

\texttt{environmental\_attributes.csv}
& 3,757
& 47
& Settlement-level hydrological, coastal, terrain and land-cover attributes,
including OS Terrain 50 measures at multiple neighbourhood scales,
Copernicus GLO-90-derived terrain attributes, CORINE-derived land-cover and
woody-cover variables, and associated quality or eligibility fields. \\

\texttt{lexical\_registry.csv}
& 24
& 11
& Frozen lexical registry containing canonical Welsh place-name elements,
project-authored concise glosses, semantic domains, preregistered roles,
source identifiers and locators, interpretation limitations and registry
version information. \\
\hline
\end{tabular}
\end{table}

\subsection{\texttt{settlements.csv}}

The file \texttt{settlements.csv} contains one row for each of the 3,757
settlement records retained in the Wales-wide analytical frame. A record
represents a unique Ordnance Survey Open Names settlement feature at a mapped
point location rather than a unique settlement name. Consequently, repeated
names are retained as separate geographic records.

The retained frame contains five settlement classes:
\texttt{Village}, \texttt{Hamlet}, \texttt{Suburban Area},
\texttt{Other Settlement} and \texttt{Town}. Each record includes the
project-level \texttt{templar\_id}, the upstream
\texttt{os\_open\_names\_id}, source name fields and associated language
metadata, the derived analytical name used by the lexical-processing pipeline,
and fields identifying the source of that analytical name and its language
status.

The analytical-name construction preserves an explicitly Welsh-labelled source
name where one is available under the frozen name-selection rule; otherwise the
primary Open Names field is retained and the analytical language status is
recorded as unresolved. The dataset therefore distinguishes absence of an
explicit language label from evidence that a name belongs to a particular
language.

Geographic location is provided in British National Grid coordinates
(EPSG:27700), together with latitude and longitude derived through the documented
coordinate transformation. The table also contains grouping variables used to
describe repeated normalised names and coincident point locations. Across the
3,757 records there are 3,328 normalised analytical-name groups and 3,755
coordinate clusters. Four records form two pairs of coincident coordinates;
these records are retained because they correspond to distinct upstream
settlement records.

Additional derived naming fields describe the normalised analytical-name group,
name length and token count. These attributes support reuse of the dataset for
linguistic, spatial and data-quality analyses without requiring users to repeat
the frozen analytical-name construction.

\subsection{\texttt{lexical\_detections.csv}}

The file \texttt{lexical\_detections.csv} contains the complete long-format
output of the frozen ETDE v1 lexical-detection procedure applied to the
analytical names of all 3,757 settlements. Each row represents one deterministic
lexical detection and is linked to the corresponding settlement by
\texttt{templar\_id}.

The table contains 1,350 detection records distributed across 1,294 settlements.
Of these detections, 378 are exact-token matches and 972 are prefix-token
matches under the frozen ETDE v1 matching rule. A total of 2,463 settlements
have no detection from the 24-element registry, 1,238 settlements have exactly
one detection and 56 have two detections; no settlement has more than two
detections. Fifty-seven detections originate from analytical names selected
from a Welsh-labelled secondary name field.

For each detection, the table records the canonical lexical element, the matched
token, token index or position where defined, match type, semantic
classification, the element's preregistered analytical role where applicable,
the source of the analytical name, and the relevant ETDE and lexical-registry
versions.

The \texttt{preregistered\_role} field is a provenance annotation rather than
an inferential result. It identifies whether an element participated in one of
the prespecified analytical contrasts or had no preregistered role. The lexical
detections themselves are reproducible string-level annotations and should not
be interpreted as validated morphological analyses, language classifications or
individual-name etymologies.

\begin{figure}[htbp]
\centering
\includegraphics[width=\textwidth]{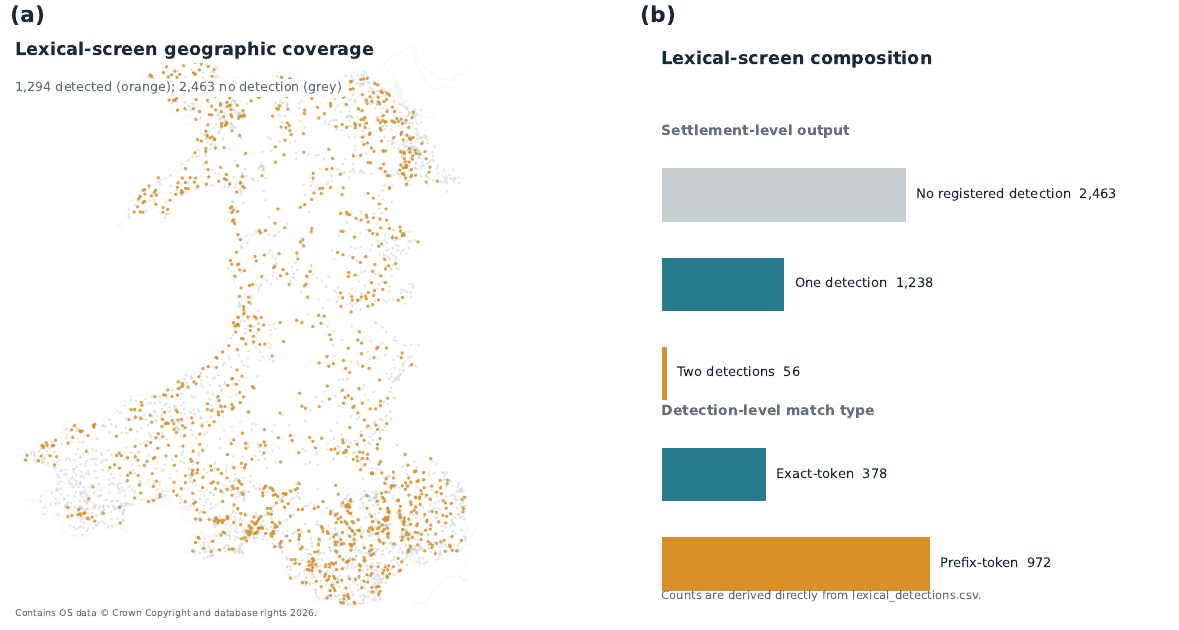}
\caption{\textbf{Geographic coverage and composition of the released lexical
resource.} \textbf{(a)} Settlements with one or more registered ETDE v1
detections are shown in orange (n = 1,294); settlements with no registered
detection are shown in grey (n = 2,463). \textbf{(b)} At settlement level,
1,238 records have one detection and 56 have two detections. At detection
level, the long-format table contains 378 exact-token and 972 prefix-token
detections. These are descriptive properties of the released lexical resource;
absence of a registered detection does not imply absence of Welsh etymology or
historical linguistic significance. Contains OS data \textcopyright\ Crown
Copyright and database rights 2026.}
\label{fig:lexical_coverage}
\end{figure}

The geographic coverage and composition of the lexical-screen output are
summarised in Fig.~\ref{fig:lexical_coverage}.

\subsection{\texttt{environmental\_attributes.csv}}

The file \texttt{environmental\_attributes.csv} contains one row for each
\texttt{templar\_id} and provides 47 settlement-level environmental and
quality-control fields. These attributes describe contemporary hydrological,
coastal, terrain and land-cover conditions derived from the source products
documented in \texttt{source\_provenance.csv}.

Hydrological fields include distance to the nearest mapped OS Open Rivers
feature and associated processing or eligibility information. Coastal context
is represented by distance to the mapped High Water Mark derived from OS
Boundary-Line.

Terrain attributes derived from OS Terrain 50 include point elevation,
neighbourhood terrain summaries and local topographic-position measures at
1-km, 2-km and 5-km neighbourhood scales, together with retained validity or
coverage information. Corresponding 2-km terrain attributes derived from
Copernicus DEM GLO-90 provide an independently produced elevation-based
measurement layer.

Land-cover fields derive from CORINE 2018 and include point-level class
information, assignment or quality indicators and woody-cover summaries at
500-m, 1-km and 2-km neighbourhood scales where the frozen coverage criteria
are satisfied. The environmental table retains missing values where the
documented coverage or eligibility criteria are not met rather than replacing
such values with zero.

The core hydrological, coastal, OS Terrain 50, Copernicus GLO-90 and point-level
CORINE fields are available for all 3,757 settlements. Woody-cover summaries
are available for 3,619 records at 500~m, 3,463 records at 1~km and 3,216
records at 2~km. Their corresponding coverage and eligibility fields should be
used when interpreting missing values.

All environmental variables are observational or deterministically derived
data attributes. Statistical coefficients, confidence intervals, significance
tests, model predictions, residuals and spatial cross-validation results from
subsequent analyses are deliberately excluded from the dataset.

\subsection{\texttt{lexical\_registry.csv}}

The file \texttt{lexical\_registry.csv} contains the frozen 24-element Welsh
place-name registry used by ETDE v1. Each row represents one canonical lexical
element and contains a concise project-authored general gloss, semantic domain,
preregistered analytical role where applicable, source identifier and locator,
and project-authored information describing alignment and limitations on
interpretation.

The registry is designed as a documented lexical resource rather than a
dictionary extract. Source references point users to the underlying RCAHMW and
Geiriadur Prifysgol Cymru materials, while the released glosses and
classification fields are concise TEMPLAR-authored summaries. The registry
therefore supports reproducible lexical annotation without reproducing
substantial dictionary definitions.

The canonical-element field provides the relational link between
\texttt{lexical\_detections.csv} and the registry. Every detection in the
released dataset maps to exactly one registry element.

\subsection{Supporting metadata, provenance and integrity records}

The four scientific tables are accompanied by six supporting files that define
their interpretation and reproducibility.

\texttt{data\_dictionary.csv} provides field-level metadata for every released
column, including description, data type, units, allowed categorical values,
missing-value semantics, source, derivation, spatial scale, coordinate
reference information, licensing family and validation rule.

\texttt{source\_provenance.csv} records the upstream products and lexical
sources used to construct the resource, including provider, product or source
name, version or dated snapshot where available, transformation into released
fields, licensing or data-use terms and the level of provenance evidence
retained for the source.

\texttt{README.md} describes the relational structure of the resource,
identifier conventions, analytical-name selection, lexical annotation,
environmental attributes, missingness and appropriate reuse.
\texttt{LICENSES.md} documents the field-aware licensing model and required
attributions for TEMPLAR-authored, Ordnance Survey-derived and Copernicus-derived
content.

Finally, \texttt{manifest.json} records release-level metadata, file structure,
versioning and build information, while \texttt{SHA256SUMS.txt} provides
integrity hashes for the public release files.

\section{Technical Validation}

Technical validation was designed to assess the integrity, reproducibility and
measurement consistency of the released data resource rather than to retest the
environmental hypotheses for which the dataset was originally developed. We
therefore evaluated the settlement frame, lexical annotations, relational
structure and field constraints, environmental-data coverage, agreement between
independently sourced terrain measurements, and deterministic reconstruction of
the public release. Inferential results from subsequent applications of the
dataset are not used as evidence of dataset validity.

\subsection{Settlement-frame integrity}

The settlement frame was validated against the frozen source-derived master
frame before public export. The released \texttt{settlements.csv} contains
3,757 records and 3,757 unique \texttt{templar\_id} values. The retained
upstream \texttt{os\_open\_names\_id} is also unique across the released
frame, providing a secondary source-traceability key. Every settlement has a
non-missing analytical name and mapped location, and all records belong to one
of the five prespecified settlement classes used during frame construction.

Validation retained repeated names and coincident locations rather than treating
either as automatic duplication errors. The 3,757 settlement records correspond
to 3,328 normalised analytical-name groups and 3,755 coordinate clusters.
Inspection of the latter identified two coincident-coordinate pairs, comprising
four records in total. These records have distinct upstream settlement
identifiers and were therefore retained as separate observations. The released
record count should consequently be interpreted as the number of retained
settlement features rather than the number of unique names or unique coordinate
locations.

The analytical-name fields were additionally checked against the frozen
name-selection rule. All 3,757 records resolve to exactly one analytical name:
3,618 use \texttt{name1}, including 203 records explicitly labelled as Welsh,
whereas 139 use a Welsh-labelled \texttt{name2}. The remaining 3,415
\texttt{name1}-derived analytical names have unresolved language status rather
than an inferred non-Welsh classification. These checks ensure that the
released analytical-name field and its provenance fields reproduce the
predefined selection procedure without missing or multiply assigned records.

Coordinate and categorical fields were subjected to release-level validation
for completeness, expected coordinate reference system, permitted settlement
classes and consistency of derived grouping fields. No settlement record failed
the required identifier, name, location or settlement-class checks in the
frozen v1.0.0 release.

\subsection{Lexical-detection validation}

The lexical-detection layer was validated at both the dataset and implementation
levels. The frozen ETDE v1 procedure was reapplied to the analytical names of
all 3,757 settlements, and the resulting records were compared with the released
\texttt{lexical\_detections.csv}. The reconstruction produced 1,350 detection
records across 1,294 settlements, reproducing the released table exactly at the
level of settlement--element detections and their associated matching metadata.

Internal consistency checks confirmed the expected distribution of detection
types. Of the 1,350 detections, 378 are exact-token matches and 972 are
prefix-token matches. At settlement level, 2,463 records contain no registered
detection, 1,238 contain one detection and 56 contain two detections; no
settlement contains more than two detections. Fifty-seven detections originate
from analytical names selected from the Welsh-labelled \texttt{name2} field.
Together, these counts account for the complete settlement frame and the
complete released detection table without orphaned or multiply unresolved
records.

The detector was also checked against fixed element-level counts preserved in
the frozen analytical record. The four terrain-related elements reproduced
exactly: \textit{bryn} (84 detections), \textit{mynydd} (17),
\textit{cwm} (100) and \textit{pant} (39). Under the corresponding frozen
registry classifications, these comprise 101 high-terrain and 139 low-terrain
records, with no record assigned to both polarity groups. These quantities are
used here solely as deterministic regression targets for validating the lexical
export and should not be interpreted as evidence for an association between
place names and terrain.

Validation additionally enforced referential integrity between the detection
table and the frozen 24-element lexical registry \cite{RCAHMWPlaceNames,GPCOnline}. Every released
\texttt{canonical\_element} resolves to exactly one registry entry, every
detection is linked to a valid \texttt{templar\_id}, and each stored
\texttt{match\_type} belongs to the exact- or prefix-token categories defined
by ETDE v1. Detector and registry version fields are retained with the released
records so that each annotation can be traced to the rules and lexical resource
under which it was generated.

The implementation-level validation suite also checks the frozen text
normalisation and matching behaviour, including lowercasing, Unicode NFKD
decomposition and accent removal, conversion of hyphens and apostrophes to
token boundaries, removal of characters outside \texttt{a--z}, and preservation
of exact- versus prefix-token match types. These checks ensure that the public
annotation table can be deterministically reconstructed from the released
settlement names, lexical registry and ETDE implementation.

These validation procedures establish computational reproducibility and
internal consistency of the lexical annotation layer, but they do not constitute
independent etymological validation \cite{Owen2015,Parry2023}. ETDE is a deterministic lexical screen:
a reproduced detection demonstrates that a registered string form satisfies the
frozen matching rule in the selected analytical name. It does not establish
that the detected form is the historically correct morphological component of
that individual place name, determine the language or historical period in
which the name was coined, or validate the environmental meaning of an
individual settlement name. The lexical registry and its source-alignment
metadata are therefore released alongside the detections so that these
interpretative limits remain explicit for downstream users.

\subsection{Relational and schema validation}

The released tables were validated as a relational data resource rather than as
independent flat files. The \texttt{templar\_id} field is unique and non-missing
for all 3,757 records in \texttt{settlements.csv}, and the same set of
identifiers occurs exactly once in
\texttt{environmental\_attributes.csv}. The resulting relationship between
these two tables is therefore one-to-one. In
\texttt{lexical\_detections.csv}, \texttt{templar\_id} defines a one-to-many
relationship with the settlement frame: every detection resolves to a valid
settlement identifier, whereas settlements without a registered lexical
detection legitimately have no corresponding detection row.

The second relational link connects
\texttt{lexical\_detections.csv} to \texttt{lexical\_registry.csv}. Every
canonical element occurring in the 1,350 detection records resolves to exactly
one of the 24 entries in the frozen lexical registry. Validation found no
orphaned settlement identifiers, unresolved lexical elements or duplicate
primary records in the one-to-one tables. The retained
\texttt{os\_open\_names\_id} values are also unique across the settlement
frame, providing an additional source-traceability check independent of the
project-level identifier.

Schema validation was performed against the expected structure of each released
table. The frozen v1.0.0 release contains 3,757 rows and 23 columns in
\texttt{settlements.csv}, 1,350 rows and 13 columns in
\texttt{lexical\_detections.csv}, 3,757 rows and 47 columns in
\texttt{environmental\_attributes.csv}, and 24 rows and 11 columns in
\texttt{lexical\_registry.csv}. Validation checks enforce required column
presence, expected data types and categorical domains, identifier constraints
and table-specific structural rules. These checks are applied during generation
of the public export rather than being performed only after release packaging.

Field-level interpretation is defined in the accompanying
\texttt{data\_dictionary.csv}, which documents 94 released fields across the
four scientific tables. Each dictionary record specifies the field definition
and, where applicable, data type, units, permitted values, source or derivation,
spatial scale, coordinate reference information and validation requirements.
The dictionary is therefore treated as part of the released schema rather than
as optional descriptive documentation.

Missingness was also validated at field level. Thirteen released fields are
nullable under the frozen schema and have explicit field-specific missing-value
semantics in the data dictionary. For these fields, a missing value can
represent a documented condition such as unavailable source metadata or failure
to satisfy an environmental coverage or eligibility criterion. For fields
defined as non-nullable, a blank or missing value is treated as an export defect
rather than as a substantive data category. This distinction prevents
structural missingness from being silently conflated with measured values such
as zero.

The export validation additionally checks for material that is not permitted in
the public dataset. Raw provider geometries and raster data, local filesystem
paths, credentials, model coefficients, fitted values, residuals,
cross-validation outputs and other analysis-specific results are excluded from
the four scientific tables. These structural and content checks passed for the
frozen v1.0.0 release.

\subsection{Environmental coverage and missingness}

Environmental-field coverage was evaluated against the complete 3,757-record
settlement frame and against the field-specific coverage and eligibility rules
used during dataset construction. The purpose of these checks was to distinguish
genuine absence of an eligible environmental measurement from missing values
introduced by an incomplete export or failed table join.

The principal hydrological, coastal and terrain attributes are complete across
the released settlement frame. Distance to the nearest mapped OS Open Rivers
feature and distance to the OS Boundary-Line High Water Mark are available for
all 3,757 settlements. The OS Terrain 50-derived point elevation,
neighbourhood terrain-reference and local terrain-position fields at the
released 1-km, 2-km and 5-km scales likewise have complete settlement-level
coverage. The corresponding Copernicus DEM GLO-90-derived 2-km terrain fields
are available for all 3,757 records. These completeness checks also verify that
each environmental record resolves to the same settlement frame through
\texttt{templar\_id}, so absence of an environmental value cannot be attributed
to an unmatched settlement record.

Point-level CORINE 2018 land-cover information and its associated assignment
and quality fields are also represented across the complete settlement frame.
Neighbourhood woody-cover variables, however, were intentionally retained only
where the frozen coverage and eligibility criteria were satisfied. Valid
woody-cover fractions are available for 3,619 of 3,757 settlements at 500~m,
3,463 at 1~km and 3,216 at 2~km, corresponding to 96.3\%, 92.2\% and 85.6\%
of the settlement frame, respectively. Coverage therefore decreases as the
neighbourhood scale increases, as a larger spatial support must satisfy the
retained eligibility requirements.

\begin{table}[htbp]
\centering
\caption{Coverage of the principal environmental attribute groups in the
released settlement frame. Coverage refers to records with an eligible released
measurement and should be interpreted together with the corresponding quality,
coverage and eligibility fields.}
\label{tab:environmental-coverage}
\begin{tabular}{p{0.60\textwidth} r r}
\hline
\textbf{Attribute group} &
\textbf{Available records} &
\textbf{Coverage (\%)} \\
\hline
Nearest-river distance (OS Open Rivers) & 3,757 / 3,757 & 100.0 \\
Coastal distance (OS Boundary-Line HWM) & 3,757 / 3,757 & 100.0 \\
OS Terrain 50 attributes & 3,757 / 3,757 & 100.0 \\
Copernicus GLO-90 2-km terrain attributes & 3,757 / 3,757 & 100.0 \\
CORINE point-level land-cover attributes & 3,757 / 3,757 & 100.0 \\
Woody-cover fraction, 500~m & 3,619 / 3,757 & 96.3 \\
Woody-cover fraction, 1~km & 3,463 / 3,757 & 92.2 \\
Woody-cover fraction, 2~km & 3,216 / 3,757 & 85.6 \\
\hline
\end{tabular}
\end{table}

Missing woody-cover values are therefore an expected property of the released
resource rather than evidence of failed processing. The corresponding coverage
and eligibility fields preserve the reason that a neighbourhood-level value is
or is not available. Missing fractions were not imputed and were not converted
to zero: a zero value represents an eligible neighbourhood in which the
specified cover fraction is measured as zero, whereas a missing value indicates
that no eligible measurement is released under the frozen construction rule.
This distinction is encoded in the data dictionary and should be preserved in
downstream analyses.

Range and consistency checks were additionally applied to the derived
environmental fields. Fractional land-cover variables were required to satisfy
their documented numerical domains when present, and validity, coverage and
eligibility indicators were checked for consistency with the presence or
absence of their associated measurements. Required environmental fields were
checked for unexpected missing values, and the final export contains no
unexplained missingness in fields defined as complete by the frozen schema.

These checks establish the completeness and internal consistency of the
released environmental attributes under their documented construction rules.
They do not establish that any particular environmental source is error-free or
that measurements derived from different products are interchangeable.
Agreement between the independently sourced OS Terrain 50 and Copernicus
GLO-90 terrain representations is evaluated separately below.

\subsection{Independent terrain-source agreement}

The terrain attributes derived from OS Terrain 50 \cite{OSTerrain502025} were
independently evaluated using Copernicus DEM GLO-90 \cite{CopernicusGLO90} as
an external measurement source. This comparison was designed to test whether
the settlement-level terrain representation is robust to the choice of
elevation product rather than to test an
environmental-toponymy hypothesis. The two terrain products have independent
production lineages and different native spatial resolutions, making agreement
between their derived settlement-level measures informative about the stability
of the terrain variables released with the dataset.

The comparison was performed for all 3,757 settlements at the 2-km analysis
scale, for which corresponding terrain-reference and local-position measures
were generated from both products. Two quantities were evaluated separately:
the mean terrain elevation defining the local terrain reference and the
settlement's elevation relative to that reference. For paired measurements
$x_i$ and $y_i$ from OS Terrain 50 and Copernicus GLO-90, respectively,
agreement was summarised using Pearson correlation and the mean absolute error
(MAE),

\begin{equation}
\mathrm{MAE} =
\frac{1}{n}\sum_{i=1}^{n}\left|x_i-y_i\right|,
\end{equation}

with the root mean squared error (RMSE) and mean signed difference additionally
retained for the terrain-reference comparison. Agreement was additionally
visualised using difference-versus-mean plots, with the mean paired difference
and 95\% limits of agreement defined as the mean difference $\pm 1.96$ standard
deviations of the paired differences.

Agreement between the independently derived 2-km terrain-reference measurements
was high across the settlement frame (Pearson $r=0.9999158$). The mean signed
difference between the two products was 1.05~m, with an MAE of 1.18~m and an
RMSE of 1.63~m. These differences are small relative to the between-settlement
variation in local terrain elevation and indicate that the neighbourhood
terrain reference is reproduced closely when calculated from the independent
elevation product.

Agreement remained strong after converting the elevation surfaces into the
local terrain-position measure used in the dataset. The OS Terrain 50- and
Copernicus GLO-90-derived 2-km local-position values had a Pearson correlation
of $r=0.9950185$ and an MAE of 2.70~m. The slightly lower agreement for local
position than for the underlying terrain reference is expected because this
quantity combines the settlement elevation and its neighbourhood terrain
reference, allowing product-specific differences in both components to
contribute to the resulting deviation.

\begin{table}[htbp]
\centering
\caption{Agreement between settlement-level terrain attributes independently
derived from OS Terrain 50 and Copernicus DEM GLO-90. All comparisons use the
complete 3,757-settlement frame at the 2-km analysis scale.}
\label{tab:terrain-source-validation}
\begin{tabular}{p{0.3\linewidth}rrrr}
\hline
\textbf{Terrain measure} &
\textbf{Pearson $r$} &
\textbf{Mean difference (m)} &
\textbf{MAE (m)} &
\textbf{RMSE (m)} \\
\hline
Terrain reference
& 0.9999158 & 1.05 & 1.18 & 1.63 \\
Local terrain position
& 0.9950185 & 0.17 & 2.70 & 4.04 \\
\hline
\end{tabular}
\end{table}

\begin{figure}[htbp]
\centering
\includegraphics[width=\textwidth]{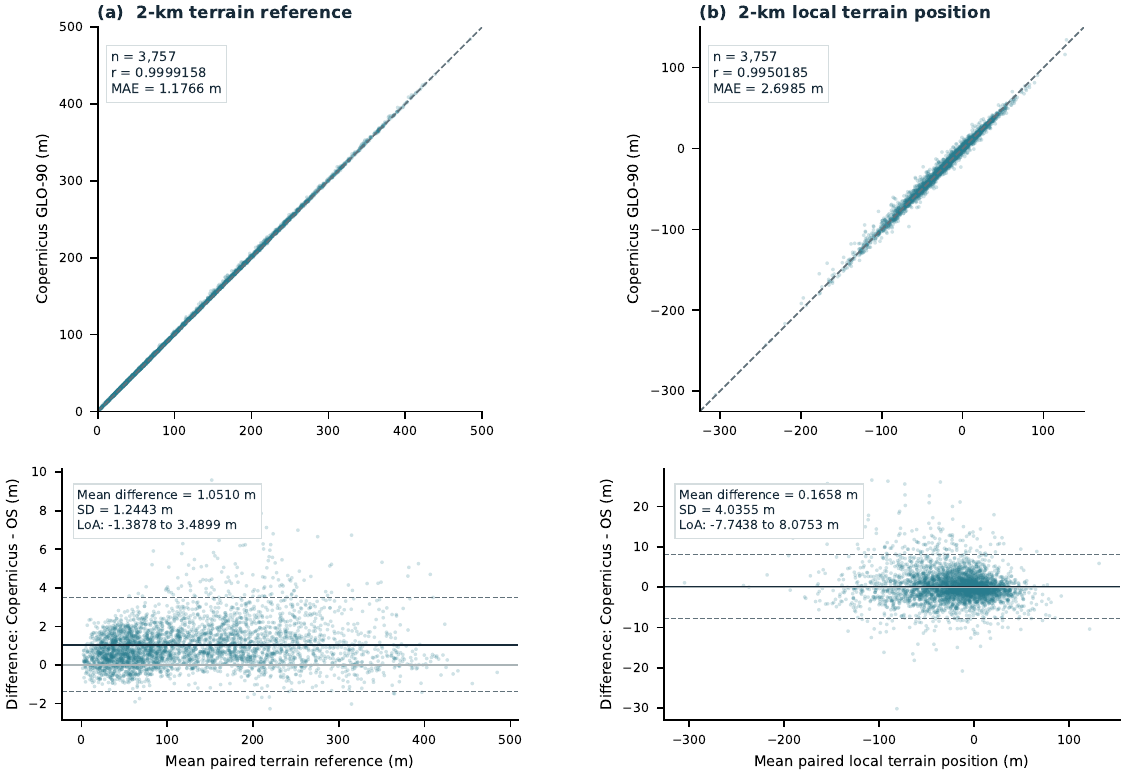}
\caption{\textbf{Independent terrain-source technical validation.} Agreement
between terrain attributes independently derived from OS Terrain 50 and
Copernicus DEM GLO-90 for all 3,757 settlements. \textbf{(a)} 2-km terrain
reference and \textbf{(b)} 2-km local terrain position. In each column, the
upper panel shows paired values with the 1:1 identity line; the lower panel
shows the Copernicus-minus-OS difference against the paired mean. Solid
horizontal lines show mean differences and dashed lines show 95\% limits of
agreement (mean difference $\pm 1.96$ SD). Units are metres. This comparison is
technical validation of independent terrain derivations and does not designate
either source as absolute ground truth.}
\label{fig:terrain_validation}
\end{figure}

Agreement between the independent terrain derivations is shown in
Fig.~\ref{fig:terrain_validation}.

The comparison is not intended to establish either elevation product as a
ground-truth representation of terrain. Differences can arise from native
resolution, surface construction, resampling and other product-specific
characteristics. Instead, the analysis evaluates whether the derived
settlement-level quantities depend strongly on a single elevation source. The
close correspondence of both the neighbourhood terrain reference and local
terrain-position measure indicates that the released terrain representation is
not an artefact of the OS Terrain 50 elevation product alone.

Both sets of terrain attributes are therefore retained in
\texttt{environmental\_attributes.csv}. This allows downstream users to select
a preferred source, reproduce the cross-source comparison or assess
source-dependent sensitivity in applications for which terrain measurement is
important. The comparison provides technical validation of the measurement
layer only and does not constitute evidence for any relationship between
terrain and the lexical content of settlement names.

\subsection{Deterministic reconstruction and release integrity}

The final stage of technical validation evaluated whether the public dataset
could be reconstructed deterministically from the frozen project inputs and
whether the resulting release preserved the approved scientific-table contents.
The four scientific tables were generated through a versioned export procedure
rather than assembled or edited manually. The same procedure also generates
the accompanying dictionary, provenance, manifest and integrity information
required for release validation.

A clean rebuild was compared with the approved v1.0.0 candidate at the
scientific-table level. The reconstructed
\texttt{settlements.csv},
\texttt{lexical\_detections.csv},
\texttt{environmental\_attributes.csv} and
\texttt{lexical\_registry.csv} reproduced the frozen contents exactly. Their
scientific-table hashes were identical to those of the approved candidate,
demonstrating that repeated execution of the export procedure did not alter
record ordering, identifiers, lexical annotations, environmental values or
schema content.

Automated release tests were then applied to the frozen dataset package. These
tests jointly check table dimensions, deterministic identifier construction,
identifier uniqueness, relational joins, permitted settlement classes,
lexical-detection totals and match types, fixed element-level regression
targets, environmental-field completeness, documented woody-cover availability,
schema constraints and exclusion of prohibited content. The final post-freeze
validation suite completed successfully, with all six release tests passing.

Release integrity is additionally supported by cryptographic checksums. The
public package contains \texttt{SHA256SUMS.txt}, which records SHA-256 hashes
for the release files, and all recorded checksums were verified after the
v1.0.0 freeze. The accompanying \texttt{manifest.json} records the dataset
version and public inventory, providing a machine-readable description of the
release against which file presence and version identity can be checked.

The release-validation procedure also tests the boundary between the reusable
dataset and its upstream or downstream analytical materials. The public
dataset contains no raw third-party vector geometries, raster products or
provider data archives, and no local filesystem paths or credentials.
Similarly, statistical coefficients, confidence intervals, hypothesis-test
results, fitted values, residuals, permutation outputs and spatial
cross-validation results are excluded. The resulting release therefore
contains the settlement-level resource, its lexical and environmental
attributes, and the metadata required to interpret and validate those records,
without incorporating analysis-specific outputs from subsequent applications.

These checks validate the integrity of the frozen v1.0.0 release rather than
implying that the dataset is immutable. Future corrections or extensions can be
issued as explicitly versioned releases, while the identifier, registry and
detector version fields allow users to determine which construction rules apply
to a particular record. The combination of deterministic export, automated
validation, manifest information and file-level checksums provides a
reproducible boundary around the dataset version described here.

\section{Usage Notes}

TEMPLAR Wales is intended as a reusable settlement-level resource for research
at the intersection of toponymy, linguistic geography, environmental history,
geographical information science and spatial data analysis. The four-table
structure allows users to work with the settlement frame, lexical annotations
and environmental measurements either jointly or independently. Appropriate
reuse nevertheless requires attention to the observational unit, the scope of
the lexical detector, source-specific environmental measurements, spatial
dependence and the documented missing-value conventions.

\subsection{Joining and selecting data records}

For analyses requiring one record per settlement,
\texttt{settlements.csv} should normally be treated as the reference table.
It can be joined one-to-one with
\texttt{environmental\_attributes.csv} using \texttt{templar\_id}. The
\texttt{lexical\_detections.csv} table has a one-to-many relationship with
the settlement frame and should therefore not be joined directly without
considering the resulting change in observational unit: a settlement with two
detections will otherwise contribute two rows, whereas a settlement with no
registered detection will contribute none.

Users requiring settlement-level lexical indicators can derive these from the
long-format detection table according to their research question, while
retaining settlements with no detections explicitly where appropriate.
Canonical lexical elements can be joined to
\texttt{lexical\_registry.csv} for semantic, provenance and source-alignment
metadata. The project-level \texttt{templar\_id} is recommended for joins
within TEMPLAR Wales; \texttt{os\_open\_names\_id} is retained principally
for source traceability.

The accompanying \texttt{data\_dictionary.csv} should be consulted before
selecting fields programmatically. In particular, units, spatial scales,
categorical domains, nullable status and missing-value semantics are defined at
field level rather than assumed to be common across variables.

\subsection{Settlement records and place names}

The 3,757 observations should not be interpreted as 3,757 unique place names.
They are retained OS Open Names settlement records at mapped locations. The
resource contains 3,328 normalised analytical-name groups and 3,755 coordinate
clusters, and repeated names or coincident locations can therefore occur
legitimately.

The appropriate unit of analysis depends on the reuse question. Studies of
settlement locations can generally retain the source-record frame, whereas
studies concerned with lexical diversity, unique written forms or name
frequency may need to group or deduplicate records using the supplied
normalised-name information. Such aggregation changes the observational unit
and should be reported explicitly.

Likewise, the analytical name is a project-defined processing field rather than
a replacement for the available source naming information. It identifies the
single name selected under the frozen rule for deterministic lexical screening.
Users interested in bilingual naming, name variants or individual historical
toponymy should consult the original naming and language-status fields and,
where necessary, appropriate historical or linguistic sources rather than
treating the analytical name as a complete representation of the settlement's
naming history \cite{Owen2015,Parry2023}.

\subsection{Interpreting lexical detections}

The lexical detections are reproducible string-level annotations produced by
ETDE v1 against a frozen 24-element registry. They are suitable for applications
that require a transparent and repeatable lexical screen, including
settlement-level lexical grouping, spatial distributions of registered forms,
comparison with environmental or cultural attributes, and methodological work
on rule-based toponym detection.

A detection should not, however, be treated as a verified etymology. Exact- and
prefix-token matches establish that a registered string form satisfies the
documented ETDE rule in the selected analytical name; they do not independently
demonstrate morphological identity, historical sense, linguistic origin or the
circumstances in which an individual settlement was named. Prefix matches in
particular should remain distinguishable from exact-token matches when the
research question is sensitive to lexical specificity
\cite{Owen2015,Parry2023}.

The registry's semantic classifications and concise glosses should similarly
be interpreted as reproducible project annotations linked to documented lexical
sources, not as exhaustive dictionary definitions. For research focused on
individual place-name histories, the source locators supplied in
\texttt{lexical\_registry.csv} provide starting points for further specialist
verification.

The absence of a detection is also specific to the released registry and
detector. A settlement with no row in \texttt{lexical\_detections.csv} has no
match to one of the 24 registered elements under ETDE v1; it should not be
described as lacking environmental, Welsh or otherwise meaningful lexical
content.

\subsection{Language-status information}

Language metadata should be interpreted conservatively. The frozen
analytical-name procedure distinguishes names carrying an explicit Welsh source
label from names for which the relevant language status is unresolved. An
unresolved status is not equivalent to a classification as English or
non-Welsh.

Consequently, the language-status fields should not be used directly to estimate
the prevalence of Welsh versus English settlement names without an additional,
purpose-designed language-classification procedure. Researchers interested in
language contact, bilingual naming or linguistic change should retain this
distinction and may wish to enrich the dataset with independently validated
historical and linguistic information \cite{Owen2015,Parry2023}.

\subsection{Using environmental attributes}

Environmental attributes describe contemporary or product-specific geographic
conditions around the released settlement locations. They should not be
interpreted automatically as reconstructions of the landscape at the historical
time when a place name was formed. This distinction is particularly important
for land cover, river configuration and other environmental characteristics
that may change through time.

The terrain variables are supplied at multiple neighbourhood scales because
local topographic context depends on the spatial support over which it is
defined. Users should select the 1-km, 2-km or 5-km representation according
to their research question rather than treating the scales as interchangeable
replicates. The parallel OS Terrain 50 and Copernicus GLO-90 2-km fields also
permit source-sensitivity checks where terrain measurement is central to an
analysis.

River and coastal distances are measurements relative to the mapped features
represented by their respective source products. They should not be interpreted
as measures of historical channel position, hydrological connectivity,
catchment membership or past coastline configuration without additional data.

CORINE-derived land-cover classes and woody-cover fractions likewise represent
the specified contemporary land-cover product and its construction rules.
Researchers interested in historical vegetation or land use will require
independent historical environmental sources.

\subsection{Missing environmental values}

Missing values should not be replaced automatically with zero. This is
particularly important for the neighbourhood woody-cover variables, for which
availability reflects the frozen coverage and eligibility criteria. A value of
zero denotes an eligible neighbourhood with a measured fraction of zero,
whereas a missing value denotes the absence of an eligible released
measurement.

The corresponding coverage, eligibility and quality fields should therefore be
included when constructing analytical subsets. Woody-cover measurements are
available for 3,619 settlements at 500~m, 3,463 at 1~km and 3,216 at 2~km,
and analyses comparing scales should account for the fact that the eligible
record set can change with neighbourhood size. Field-specific missing-value
semantics for all nullable variables are provided in
\texttt{data\_dictionary.csv}.

\subsection{Spatial analysis and statistical reuse}

Settlement observations are geographically structured and should not generally
be assumed to constitute independent random samples. Nearby settlements can
share environmental conditions, regional naming traditions, settlement history
and other unmeasured characteristics. Conventional random train--test splitting
or statistical procedures that ignore spatial dependence can therefore produce
overly optimistic predictive assessments or uncertainty estimates
\cite{Legendre1993,Dormann2007review,Roberts2017}.

For predictive applications, spatially separated or geographically blocked
validation is preferable when the intended claim concerns generalisation to new
locations. For inferential applications, users should consider spatial
dependence, clustering, appropriate null models and the geographic scale of the
research question. The dataset deliberately does not prescribe a single
statistical solution because the appropriate method depends on the intended
application \cite{Dormann2007review,Roberts2017,Valavi2019}.

Repeated normalised names and the two coincident-coordinate pairs can also
create dependence structures relevant to particular analyses. The supplied
normalised-name and coordinate-cluster fields allow these relationships to be
identified and incorporated into sampling, grouping or sensitivity procedures
where required.

\subsection{Extension and linkage}

The stable relational structure is intended to support enrichment with
additional settlement-level information. Potential extensions include
historical map observations, archival name forms, alternative gazetteers,
geology, soils, hydrological or climatic variables, historical land cover,
administrative context and independently curated linguistic annotations.
Additional attributes can be linked through geographic location or appropriate
source identifiers while retaining \texttt{templar\_id} as the internal key
for the released settlement frame.

Extensions to the lexical resource should be versioned separately from the
frozen 24-element registry. Adding elements, changing normalisation rules or
altering exact- or prefix-matching behaviour can change the resulting detection
set and should therefore produce a new detector or registry version rather than
silently modifying ETDE v1 annotations.

Similarly, users combining TEMPLAR Wales with newer versions of the upstream
geospatial products should record those source versions explicitly. Differences
between derived values from different product releases should not be assumed to
represent environmental change unless the underlying product methodologies and
temporal comparability have been established.

\subsection{Citation, attribution and licensing}

Users should cite the dataset version used in their analysis and retain the
source-specific attribution requirements documented in
\texttt{LICENSES.md}. TEMPLAR-authored integration, annotations and
documentation, OS-derived fields, CORINE-derived fields and Copernicus
DEM-derived fields are accompanied by distinct provenance and licensing
information; the presence of these materials in a single relational resource
does not relicense third-party-derived content under a single project licence.

The released data should therefore be redistributed or incorporated into derived
resources together with the applicable source acknowledgements and licensing
conditions. \texttt{source\_provenance.csv} and \texttt{LICENSES.md} provide
the release-specific provenance and attribution information needed for this
purpose.

\section{Data Availability}

TEMPLAR Wales v1.0.0 is publicly available from Zenodo at
\url{https://doi.org/10.5281/zenodo.22107776} \cite{KarakusEyupoglu2026TEMPLAR}.

The deposited dataset consists of the following files:
\begin{itemize}
\item \texttt{settlements.csv}
\item \texttt{lexical\_detections.csv}
\item \texttt{environmental\_attributes.csv}
\item \texttt{lexical\_registry.csv}
\item \texttt{data\_dictionary.csv}
\item \texttt{source\_provenance.csv}
\item \texttt{README.md}
\item \texttt{LICENSES.md}
\item \texttt{manifest.json}
\item \texttt{SHA256SUMS.txt}
\end{itemize}

The separate Nature Communications reproducibility archive is available from
Zenodo at \url{https://doi.org/10.5281/zenodo.22079109} and from the clean
project reproducibility repository at
\url{https://github.com/oktaykarakus/templar-wales-reproducibility}. It is not
the public deposition of the Scientific Data dataset.

\section{Code Availability}

Code supporting the deterministic construction and validation of TEMPLAR Wales
is publicly available from the project reproducibility repository at

\url{https://github.com/oktaykarakus/templar-wales-reproducibility}.

The archived reproducibility materials are available from Zenodo at

\url{https://doi.org/10.5281/zenodo.22079109}. 

The repository includes the dataset-construction code, lexical detection
implementation, configuration and focused validation tests required to
reconstruct and verify the released data products when used with the documented
upstream source data. Raw third-party geospatial source data are not
redistributed and must be obtained from their respective providers under the
applicable terms described in the dataset provenance and licensing
documentation.

\section*{Acknowledgements}

The authors acknowledge Ordnance Survey for the OS Open Names, OS Open Rivers,
OS Terrain 50 and Boundary-Line products used in constructing the dataset, and
the European Union's Copernicus programme for CORINE Land Cover 2018 and
Copernicus DEM GLO-90. The authors also acknowledge the Royal Commission on the
Ancient and Historical Monuments of Wales (RCAHMW) Historic Place Names of
Wales resources and Geiriadur Prifysgol Cymru Online as lexical reference
sources used in documenting the place-name element registry.

\section*{Funding}

The authors received no specific funding for this work.

\section*{Author Contributions}

O.K. conceived and led the study and dataset development, developed the
methodology and computational framework, performed the investigation and data
analysis, curated and integrated the data, developed the software, conducted
validation and reproducibility testing, and wrote the original manuscript.
C.E. contributed to conceptualisation, methodology, investigation, validation
and data curation, and reviewed and edited the manuscript. Both authors
reviewed and approved the final manuscript.

\section*{Use of generative AI and AI-assisted technologies}

OpenAI ChatGPT/Codex and GitHub Copilot were used to assist with software
development, analysis-workflow support, and manuscript drafting and editing.
All AI-assisted outputs were reviewed and verified by the authors; numerical
results and bibliographic records were checked against the project evidence and
source records. All scientific decisions, interpretation and conclusions remain
the responsibility of the authors.

\section*{Competing Interests}

The authors declare no competing interests.

\end{document}